\documentclass{article}

\usepackage[nonatbib, preprint]{neurips_data_2024}

\usepackage[utf8]{inputenc} 
\usepackage[T1]{fontenc}    
\usepackage{hyperref}       
\usepackage{url}            
\usepackage{booktabs}       
\usepackage{amsfonts}       
\usepackage{nicefrac}       
\usepackage{microtype}      
\usepackage{xcolor}         
\usepackage{amsmath}
\usepackage{array}
\usepackage{url}
\usepackage{multicol}
\usepackage{multirow}
\usepackage{graphicx}
\usepackage{amssymb}
\usepackage{pifont}
\usepackage{subfig}

\title{SRFUND: A Multi-Granularity Hierarchical Structure Reconstruction Benchmark in Form Understanding}
\author{%
  \quad\textbf{Jiefeng Ma}$^{1}$
  \quad\textbf{Yan Wang}$^{1}$ 
  \quad\textbf{Chenyu Liu}$^{2}$
  \quad\textbf{Jun Du}$^{1}$\thanks{Corresponding author.}
  \quad\textbf{Yu Hu}$^{1}$\\
  \quad\textbf{Zhenrong Zhang}$^{1}$
  \quad\textbf{Pengfei Hu}$^{1}$
  \quad\textbf{Qing Wang}$^{1}$
  \quad\textbf{Jianshu Zhang}$^{2}$\\ 
  $^{1}$University of Science and Technology of China, Hefei, China \\ 
  $^{2}$iFLYTEK, Hefei, China \\
  \texttt{\{jfma, yanwangsa, zzr666, hudeyouxiang\}@mail.ustc.edu.cn}, \\
  \texttt{\{jundu, yuhu2, qingwang2\}@ustc.edu.cn, \{cyliu7, jszhang6\}@iflytek.com} \\
}

\begin{document}

\maketitle

\begin{abstract}
  Accurately identifying and organizing textual content is crucial for the automation of document processing in the field of form understanding. Existing datasets, such as FUNSD and XFUND, support entity classification and relationship prediction tasks but are typically limited to local and entity-level annotations. This limitation overlooks the hierarchically structured representation of documents, constraining comprehensive understanding of complex forms. To address this issue, we present the SRFUND, a hierarchically structured multi-task form understanding benchmark. SRFUND provides refined annotations on top of the original FUNSD and XFUND datasets, encompassing five tasks: (1) \textbf{word to text-line merging}, (2) \textbf{text-line to entity merging}, (3) \textbf{entity category classification}, (4) \textbf{item table localization}, and (5) \textbf{entity-based full-document hierarchical structure recovery}. We meticulously supplemented the original dataset with missing annotations at various levels of granularity and added detailed annotations for multi-item table regions within the forms. Additionally, we introduce global hierarchical structure dependencies for entity relation prediction tasks, surpassing traditional local key-value associations. The SRFUND dataset includes eight languages including \textit{English, Chinese, Japanese, German, French, Spanish, Italian, and Portuguese}, making it a powerful tool for cross-lingual form understanding. Extensive experimental results demonstrate that the SRFUND dataset presents new challenges and significant opportunities in handling diverse layouts and global hierarchical structures of forms, thus providing deep insights into the field of form understanding. The original dataset and implementations of baseline methods are available at \url{https://sprateam-ustc.github.io/SRFUND}.
\end{abstract}

\section{Introduction}
In the United States, billions of individuals and businesses submit tax returns annually,\footnote{\url{https://www.irs.gov/pub/irs-pdf/p55b.pdf}} and globally, hundreds of billions of parcels are distributed each year,\footnote{\url{https://www.pitneybowes.com/content/dam/pitneybowes/us/en/shipping-index/23-mktc-03596-2023_global_parcel_shipping_index_ebook-web.pdf}} most of which are accompanied by invoices and delivery notes. Although these documents vary in format, they are all considered forms, which serve as crucial information mediums widely used in global information and merchandise exchange. Compared to storage formats like camera-captured images or scanned documents, digitizing original forms into structured text aids in reducing storage space and facilitates information dissemination \cite{khan2015digitization}. Consequently, there has been a growing practical demand in recent years for understanding information within forms, including both textual content and document structures across various layouts and languages. 
With the rapid development of document processing technologies, significant progress has been made in the field of form understanding \cite{docformer, structurallm, layoutlm, graphdoc}, along with the establishment of a series of benchmark datasets \cite{sroie, funsd, cord, ephoie, xfund, sibr}. However, none of these existing datasets have established the global and hierarchical structural dependencies considering all elements at different granularity, including words, text lines, and entities within the forms.

\begin{figure}[t]
    \centering
    \subfloat[Word level]{\includegraphics[width=0.23\textwidth]{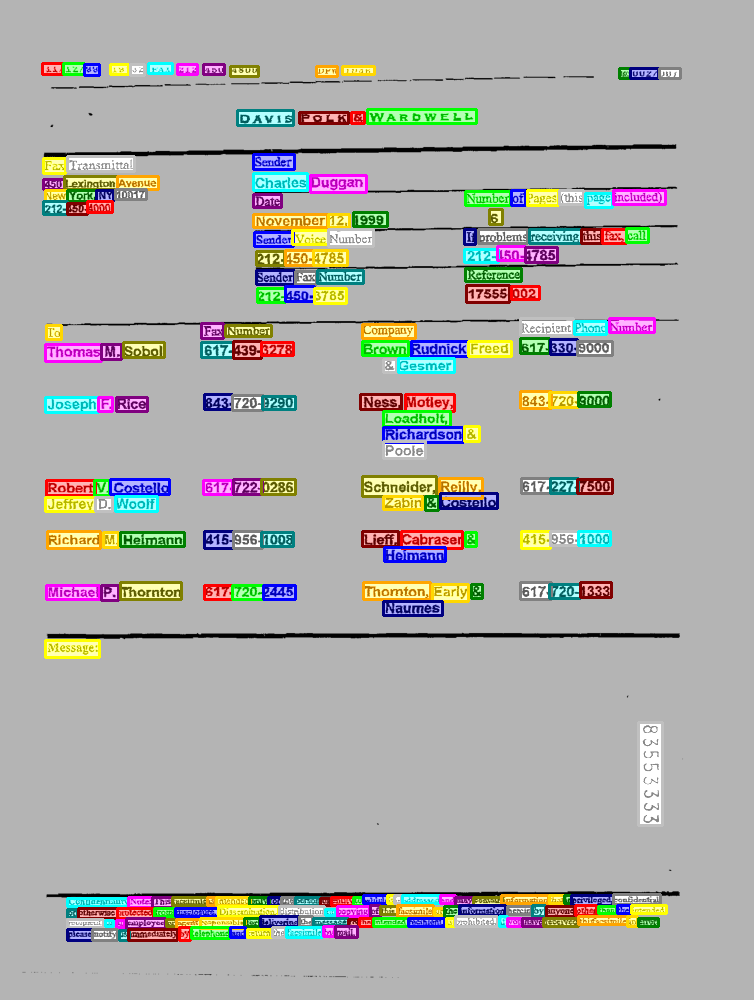}\label{fig:image1}}
    \hfill
    \subfloat[Text-line level]{\includegraphics[width=0.23\textwidth]{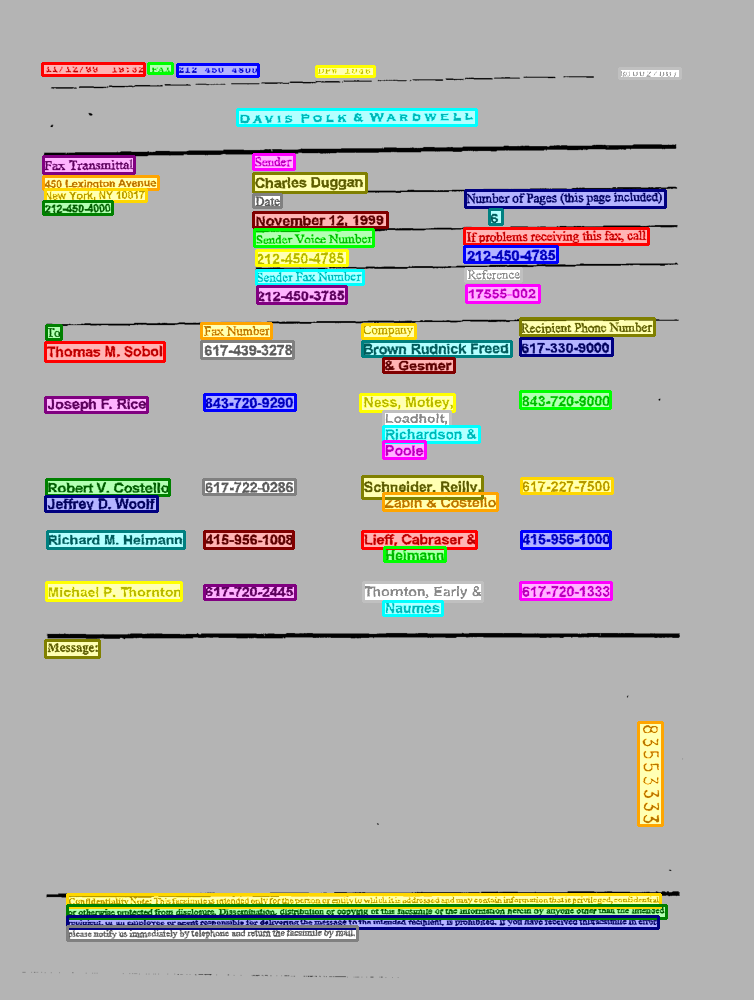}\label{fig:image2}}
    \hfill
    \subfloat[Entity level]{\includegraphics[width=0.23\textwidth]{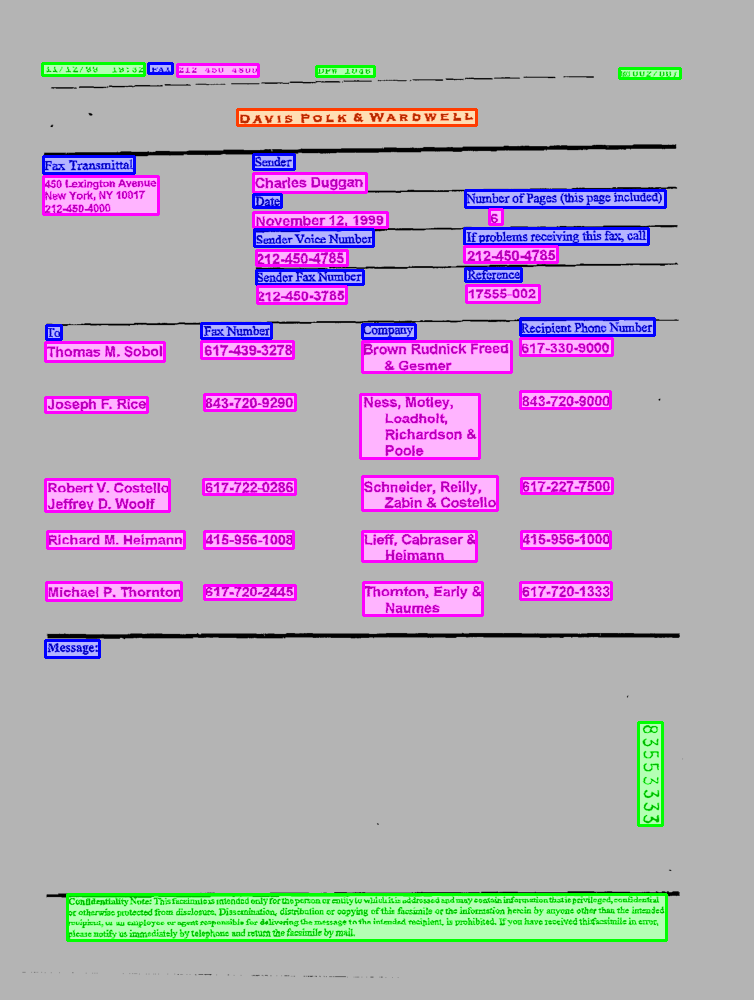}\label{fig:image3}}
    \hfill
    \subfloat[Item table level]{\includegraphics[width=0.23\textwidth]{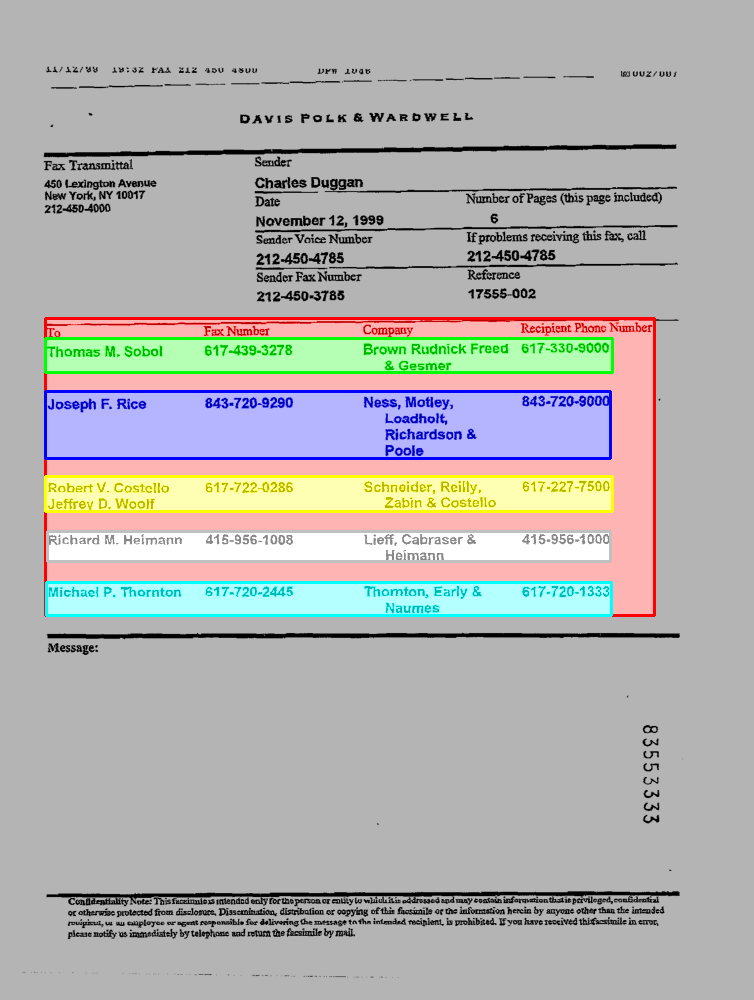}\label{fig:image4}}
    
    
    \subfloat[Overall form structure based on entities.]{\includegraphics[width=0.96\textwidth]{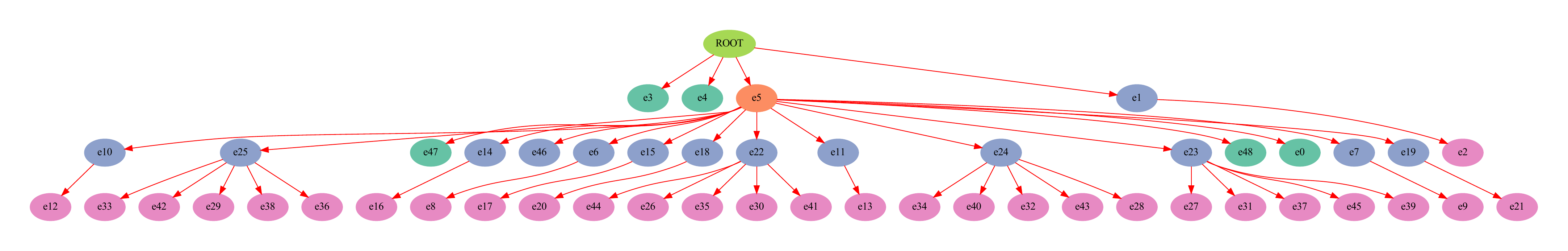}\label{fig:image5}}
    
    \caption{Multiple granularity of annotations and supported tasks on SRFUND.}
    \label{fig:SRFUND_granularity}
\end{figure}

To enhance the applicability of form understanding tasks in hierarchical structure recovery, we introduce the SRFUND, a multilingual form structure reconstruction dataset. The SRFUND dataset comprises 1,592 form images across eight languages, with each language contributing 199 images. As illustrated in Figure \ref{fig:SRFUND_granularity}, each form image is manually annotated with the locations and text contents of every word, text-line, and entity.
After identifying each independent entity, we categorize these entities into four classes including \textit{Header, Question, Answer}, and \textit{Other}, which is consistent with the FUNSD dataset definitions \cite{funsd}. Moreover, all entities in the form are annotated with their hierarchical dependencies, allowing us to reconstruct the global form structure.
For the multi-item table regions frequently found in forms, we have specifically annotated the positions of these tables, including their table headers, and grouped each line item within these tables individually. 
The refined annotations of SRFUND support the evaluation of form structure reconstruction tasks at different granularities. We conducted benchmark tests on several tasks using representative methods from three categories: vision-only, language-only, and multi-modal approaches. These tasks include (1) word to text-line merging, (2) text-line to entity merging, (3) entity category classification, (4) item table localization, and (5) entity-based full-document hierarchical structure recovery. 
Detailed experimental settings and results are presented in Sec. \ref{sec:Experiments}.




\section{Related Work}

Prior research has divided document structure tasks into two main categories: physical layout analysis and logical structure analysis \cite{haralick1994document, namboodiri2007document}. The former refers to the physical locations of various regions within a document image, while the latter aims to understand their functional roles and relationships. In this chapter, we will first review the work related to physical and logical structure analysis. Additionally, we will introduce common benchmarks widely used in form understanding tasks.

\subsection{Document physical layout analysis}
Earlier document layout analysis methods can be classified into two main categories.
Algorithms employing a bottom-up strategy start from the finest elements of the document and iteratively merge these elements based on rules or clustering algorithms to create larger and more unified regions \cite{kise1998segmentation, o1993document}. Conversely, top-down strategies begin from the entire document image and use histogram analysis or whitespace refinement methods to segment it into increasingly smaller regions \cite{ha1995recursive, zheng2003model}. With the advancement of deep learning technology, several approaches have been proposed to address the document layout analysis challenges in more complex scenarios. These approaches can generally be divided into two categories. Detection-based approaches follow the route of object detection in computer vision, treating different elements within a document as distinct detection targets \cite{li2020cross, oliveira2017fast, prasad2020cascadetabnet}. Models such as Cascade-RCNN \cite{cascadercnn}, YOLOX \cite{yolox}, and DETR \cite{detr} are used to directly predict the positions of various elements in the document images. On the other hand, methods based on instance segmentation employ frameworks used for instance segmentation in natural scene images to segment areas within documents \cite{li2019show, xu2018multi, yang2017learning}. For example, FCN \cite{fcn} or Mask R-CNN \cite{maskrcnn} is utilized to segment text-line regions or other types of areas from complex document images.

\subsection{Document logical structure analysis}
Logical structure analysis of documents focuses  on analyzing the types and relationships of document elements at a logical level, which is often built upon the results of physical layout analysis \cite{namboodiri2007document}.

In document element classification tasks, early approaches are often based on deterministic grammar rules \cite{derrien1991frame, krishnamoorthy1993syntactic, luong2012logical}. Such approaches typically require detailed rule specifications for a particular document layout and struggle to generalize to different document scenarios. Methods based on deep learning have boosted the performance on this task. Vision-only approaches for multi-class detection or segmentation do not rely on extracting the text content and position of document elements but instead employ visual models to directly locate different categories of document elements \cite{li2020cross, oliveira2017fast, yang2017learning}. Additionally, approaches based on natural language models \cite{liu2019roberta} or multi-modal language models \cite{docformer, layoutlm, graphdoc} are used to determine the types of document elements when the text content and positions of document elements are provided.

In document structure analysis tasks, early solutions employ formal grammar or logical trees to represent and arrange hierarchical relationships among elements in documents. This often requires manually designing rules tailored to the current layout. To automatically learn relationships among document elements from diverse layout data, some deep learning-based approaches have been utilized for relationship prediction tasks. DocStruct \cite{wang2020docstruct} and StrucText \cite{li2021structext} utilize a single learnable asymmetric parameter matrix for predicting asymmetric relationships between any two document elements. GraphDCM \cite{graphdcm} introduces a \textit{Merger} module containing a set of asymmetric parameter matrices, further aggregating fine-grained elements with the same category in the document into coarse-grained elements. In LayoutXLM \cite{xfund}, all possible head-tail pairs are first collected, and a bi-affine classifier \cite{biaffine} is used to determine if a relationship exists between them. GeoLayoutLM \cite{luo2023geolayoutlm} proposes a relationship classification head composed of bi-linear layers and lightweight transformers, further enhancing the performance of element relationship classification tasks.

\subsection{Form understanding benchmarks}

The development process of form understanding tasks is closely related to relevant benchmarks. SROIE \cite{sroie} comprises 973 scanned English receipts, each annotated with line-level texts, corresponding bounding boxes, and a structured extraction target with four predefined field types. CORD \cite{cord} is another receipt dataset collected from various sources, including shops and restaurants, containing 1,000 receipt images from Indonesia. Compared to SROIE, the CORD dataset includes annotations at the word and entity levels with richer extraction field types. It also provides classification attributes for key-value pairs and group information within the same item, enabling the recovery of local relationships between different entities. EPHOIE \cite{ephoie} consists of 1,494 examination paper headers collected from Chinese school exams, annotated with ten types of entities. It offers annotations for text-line-level positions and contents, as well as classification attributes for key-value pairs. However, all the aforementioned datasets are provided under specific form categories, lacking diversity in form types.
The FUNSD \cite{funsd} dataset contains 199 noisy scanned English documents, along with annotations at the word and entity levels. It categorizes all entities within forms into four classes: \textit{Header}, \textit{Question}, \textit{Answer}, and \textit{Other}, and provides local relationships between entities, supporting entity labeling and entity linking tasks. XFUND \cite{xfund} is an extension of FUNSD in multiple languages, collecting additional forms in seven languages and providing similar annotations as FUNSD. It is worth noting that XFUND suffers from some entity definition confusion, where different text lines that should belong to the same entity are split and labeled as distinct entities.
The SIBR dataset \cite{sibr} is a publicly available dataset designed for visual information extraction, comprising 1,000 form images, including 600 Chinese invoices, 300 English bills of entry, and 100 bilingual receipts. It offers text-line-level position and content information, along with two types of links to represent document element relationships: one for linking different text lines within the same entity and another for indicating relationships between different entities. However, these datasets lack uniform granularity in comprehensive annotations for words, text lines, and entities. In addition, they focus on local key-value relationships and ignore the nested relationships between elements of different hierarchies in the document, resulting in incomplete representation of form information.

\section{SRFUND benchmark}

\subsection{Data collection and annotation}\label{subsec:Data collection}

The objective of SRFUND is to advance the development of form understanding and structured reconstruction tasks.
Among existing form datasets, FUNSD and XFUND are two prominent works that have made outstanding contributions to the establishment of typical form understanding tasks and the extension to multilingual data scenarios, respectively. Our dataset, SRFUND, is built upon these two representative datasets, encompassing all document images from both datasets. SRFUND comprises 1,592 form images across eight languages, with each language contributing 199 images.

Upon careful analysis of the existing annotations in FUNSD and XFUND, we found inconsistencies in the granularity of annotations between the two datasets. The FUNSD dataset encapsulates complete semantic information of entity elements, regardless of whether this information is distributed across single or multiple lines, while the XFUND dataset includes cases where consecutive semantic multi-line text is independently counted as separate entities without connectivity between these text lines. Fortunately, both FUNSD and XFUND datasets cover word-level textual contents and positions. Leveraging the word-level annotation information from the original datasets, we meticulously followed several procedures to ensure a rigorous construction process for the dataset:
(1) Adjust inaccurate word-level bounding boxes and supplement missing textual information.
(2) Aggregate consecutive words with continuous semantics into one text-line and annotate the corresponding rectangular bounding box. It's noteworthy that separate values with bullet points, keys and values in key-value pairs are considered different text lines, even when they are visually connected to each other.
(3) For entities composed of consecutive semantic multi-line text, we annotate the polygonal bounding box of the entity.
(4) Based on the roles of entities in the current form, modify or assign correct categories to different entities. 
When the IOU between the annotation box of an entity and the existing entity box from the original datasets (i.e. FUNSD or XFUND) is greater than 0.8, the original label of the entity is adopted as the initial label of the current entity.
(5) Determine the location of item tables and annotate the headers and each individual row item within the tables.
(6) For separate entities with linkage relationships, annotate the relationships between these entities (unidirectional or bidirectional).

To ensure the accuracy of the annotations, all annotated results underwent at least three rounds of cross-checking, with any disputed annotations resolved by domain experts in the field of document processing. Given the multilingual coverage within the SRFUND dataset, a commercial translation engine was employed during the annotation process to translate form images into the native languages of annotators, providing semantic context for reference. The collection, annotation, and refinement processes of the dataset collectively consumed approximately 6,000 person-hours.

\begin{table}[h]
\caption{Comparison with existing form understanding datasets.}
\resizebox{\textwidth}{!}{
\begin{tabular}{l
>{\centering\arraybackslash}m{1.5cm}
>{\centering\arraybackslash}m{1.3cm}
>{\centering\arraybackslash}m{1.3cm}
>{\centering\arraybackslash}m{1.0cm}
>{\centering\arraybackslash}m{1.8cm}
>{\centering\arraybackslash}m{2.5cm}
>{\centering\arraybackslash}m{1.2cm}
>{\centering\arraybackslash}m{1.6cm}}
\toprule
\multirow{2}{*}{Dataset} & \multicolumn{5}{c}{Supported Tasks}  & \multicolumn{3}{c}{Statistics}  \\ \cmidrule(r{0.4em}){2-6} \cmidrule(l{0.4em}){7-9} 
 & Word to Text-line & Text-line to Entity & Entity Labeling & Item Table & Structure Recovery & Language & Images & Avg. Form Tree Depth \\ 
\midrule
SROIE \cite{sroie} &  \ding{55}  &  \ding{55}  &  \checkmark & \ding{55} &  -  & EN & 1,000 & - \\
\midrule
CORD \cite{cord} &  \checkmark  &  \checkmark  &  \checkmark & \ding{55}  &  Local  & IND & 1,000 & 1.173 \\ 
\midrule
EPHOIE \cite{ephoie} &  \ding{55}  &  \ding{55}  &  \checkmark & \ding{55} &  Local  & ZH & 1,494 & 1.115 \\ 
\midrule
SIBR \cite{sibr} &  \ding{55}  &  \checkmark  &  \checkmark & \ding{55} &  Local  & ZH, EN & 1,000 & 1.515 \\ 
\midrule
FUNSD \cite{funsd} &  \ding{55}  &  \ding{55}  &  \checkmark & \ding{55} &  Local  & EN & 199 & 1.570 \\
\midrule
XFUND \cite{xfund} &  \ding{55}  &  \ding{55}  &  \checkmark & \ding{55}  &  Local  & ZH, JA, ES, FR, IT, DE, PT & 1,393 & 1.699 \\
\midrule
SRFUND (Ours) &  \checkmark  &  \checkmark  &  \checkmark & \checkmark  &  Global  & EN, ZH, JA, ES, FR, IT, DE, PT  & 1,592 & 3.049 \\
\bottomrule
\end{tabular}
}
\label{tab:SRFUND_compare}
\end{table}

\subsection{Dataset analysis}

\textbf{Supported tasks}: 
As illustrated in Figure \ref{fig:SRFUND_granularity} and Table \ref{tab:SRFUND_compare}, SRFUND covers annotations at different levels, enabling the dataset to support a wider range of tasks than all existing form understanding datasets. It is noteworthy that previous datasets focused solely on the structural relationships between local document entities, whereas we meticulously annotated the logical relationships among all entities. This makes SRFUND the first dataset supporting the task of structure recovery for each entity at a global level. Furthermore, with the complement of item tables and their constituent items, SRFUND is also the first dataset supporting the localization of item tables within forms. The recovery of global structures and the localization of item tables require models to possess a strong understanding of entities within forms, posing significant challenges to form understanding tasks.

\begin{table}[h]
\caption{Statistics of different granularity categories of SRFUND. \textit{Total}, \textit{Same}, \textit{New} refer to the total amount of a class in SRFUND, the same amount as in the original dataset (FUNSD/XFUND), and the amount added or modified, respectively.}
\centering
\resizebox{0.7\textwidth}{!}{
\centering
\begin{tabular}{lcccccc}
\toprule
 & Words & Lines & Entities & Tables & Table Items & Links \\
\midrule
Total & 529,711 & 112,662 & 96,824 & 591 & 1,954 & 122,594 \\ 
\midrule
Same & 529,372 & 0 & 81,462 & 0 & 0 & 47,113 \\  
\midrule
New & 339 & 112,662 & 15,362 & 591 & 1,954 & 75,481 \\ 
\bottomrule
\end{tabular}
}
\label{tab:SRFUND_fix_statistic}
\end{table}

\textbf{Statistical metrics}: 
(a) As shown in Table \ref{tab:SRFUND_compare}, the SRFUND dataset includes annotations in eight different languages, making it more diverse than existing datasets and addressing form understanding needs across various languages. In terms of document hierarchy, we constructed global entity relationships, resulting in an average tree depth of 3.049, which surpasses previous datasets significantly. 
(b) As depicted in Table \ref{tab:SRFUND_fix_statistic}, the SRFUND dataset introduces a substantial number of annotations across different hierarchical levels in addition to the original data annotations. We modified 339 bounding boxes or texts at the word level and provided annotations for 112,662 text lines. Our dataset contains a total of 96,824 entities, out of which 15,362 are newly added or modified. Additionally, we provided detailed annotations for item tables in documents, totaling 591 item tables and 1,954 item contents within them. Furthermore, we added 75,481 entity links, resulting in a total of 122,594 entity links. 

\section{Experiments}\label{sec:Experiments}

\begin{figure}[h]
    \centering
    \includegraphics[width=\textwidth]{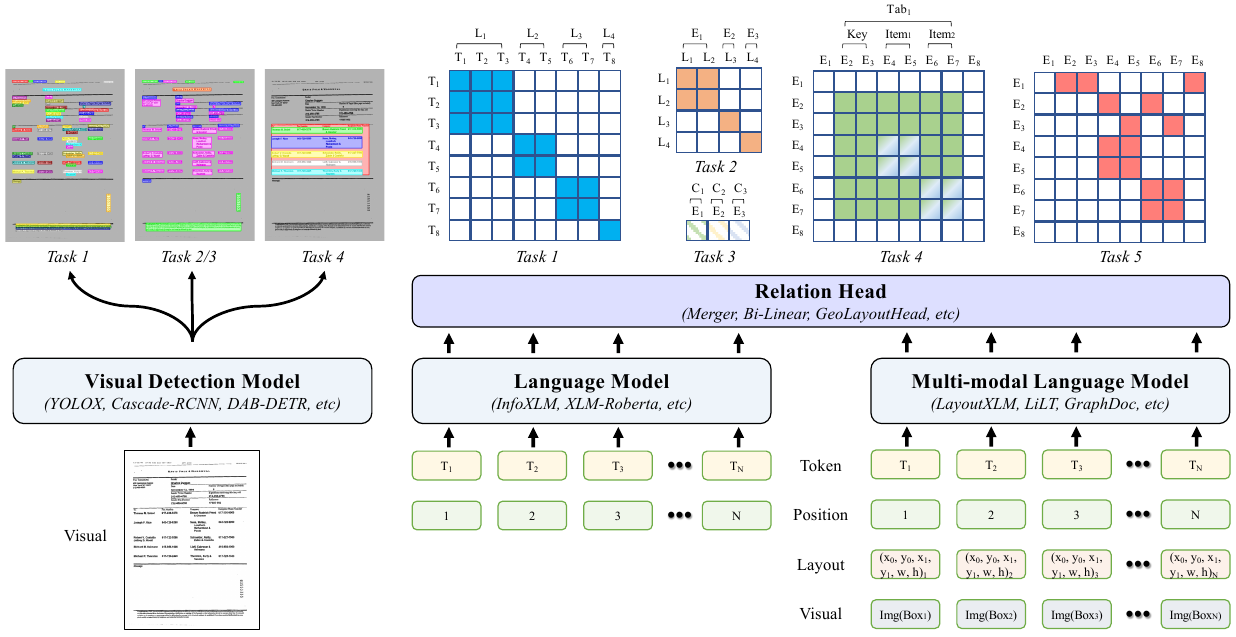}
    \caption{Models with varied modalities used for evaluating on the SRFUND benchmark. }
    \label{fig:SRFUND_exp_models}
\end{figure}

\begin{figure}[b]
    \centering
    \subfloat[Correct detection examples]{\includegraphics[width=0.498\textwidth]{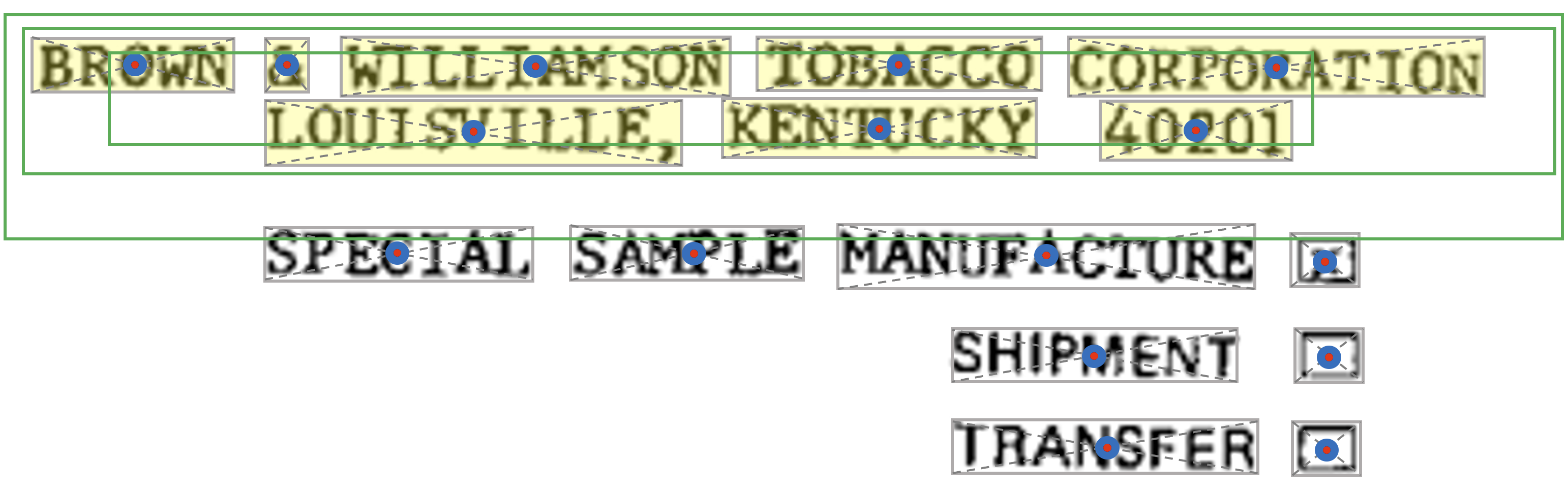}\label{fig:detect_explain_1}}
    \hfill
    \subfloat[Incorrect detection examples]{\includegraphics[width=0.498\textwidth]{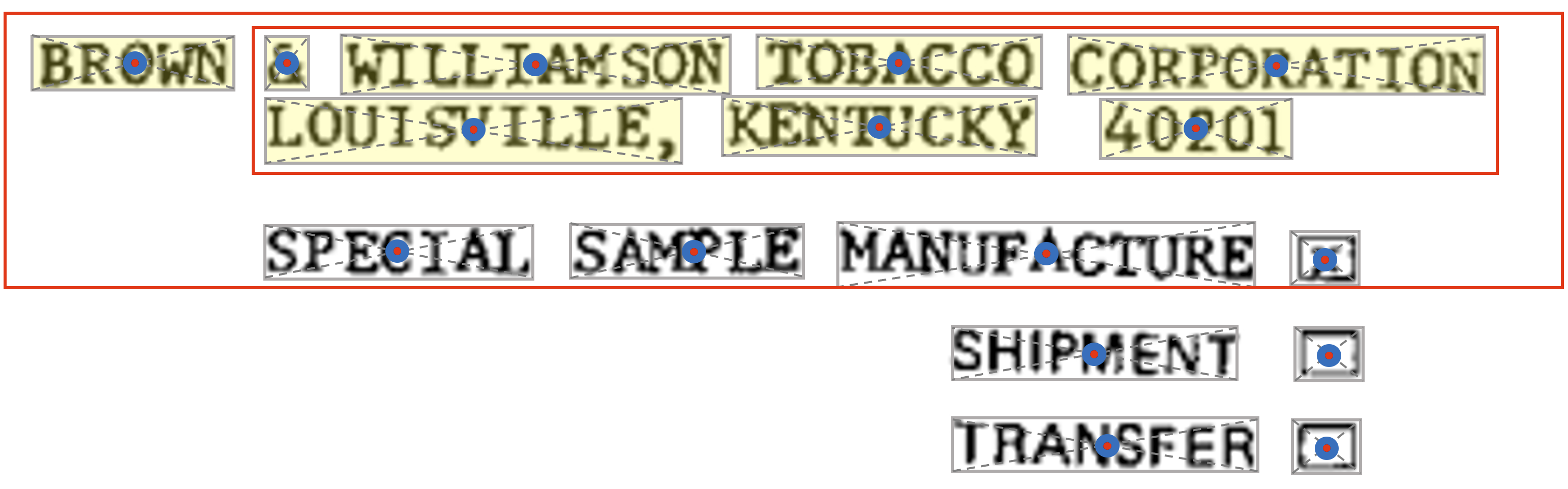}\label{fig:detect_explain_2}}

    \caption{Visualization of correct (the green boxes) and incorrect (the red boxes) bounding box predictions to capture the \textit{Header} entity (texts with yellow background). Bounding box must include exactly the word-level centers that lie within the ground truth annotation. Note: in Figure \ref{fig:detect_explain_1}, only one of the predictions would be considered correct if all three boxes were predicted.}
    \label{fig:detect_metric}
\end{figure}

To comprehensively assess the SRFUND dataset, we conducted benchmark tests using models across three different modalities: language models based on pure text input, detection models based on purely visual inputs, and document pre-trained language models that utilize multi-modal inputs. 
We performed experimental analyses on five tasks, detailed as follows: (1) word to text-line merging, (2) text-line to entity merging, (3) entity category classification, (4) item table localization, and (5) entity-based full-document hierarchical structure recovery. 

\subsection{Setting}


As illustrated in Figure \ref{fig:SRFUND_exp_models}, the vision-only models rely on the document image as the input, and tasks 1 to 4 can be regarded as the text-line detection task, the entity detection task, the multi-class entity detection task and the line item table area detection task respectively.
We selected three distinct types of visual detection models for comparison: YOLOX \cite{yolox}, a single-stage detector based on the YOLO architecture; Cascade-RCNN \cite{cascadercnn}, a multi-stage detector based on the RCNN framework; and DAB-DETR \cite{dabdetr}, an improved version of the DETR model with faster convergence speed and detection accuracy. All three detection models utilize a ResNet-50 \cite{resnet} backbone pre-trained on ImageNet \cite{imagenet}. In the training stage, we follow the original configuration adopted in the mmdetection \cite{mmdetection}. During testing, we only preserve predicted boxes with a score threshold larger than 0.3 and adopt the non-maximum suppression algorithm for further filtering. 
Since the task is framed as a coarse-to-fine merging task, we use the standard F1 score as the main evaluation metric. Unlike the common practice in object detection, where true positives are determined by thresholding the Intersection-over-Union, we use a different criterion tailored to evaluate the usefulness of detections for text read-out. Inspired by the CLEval metric \cite{cleval} used in text detection, we measure whether the predicted area contains nothing but the related word-level box centers as visualized in Figure \ref{fig:detect_metric}.

The text-only models rely on the word texts as the input, and the multi-modal models rely on the texts, two-dimensional coordinates, and form images as the input. 
Given the multilingual nature of the SRFUND dataset, we employed several multilingual language models and document pre-trained language models that support multilingual inputs. The language models include InfoXLM \cite{infoxlm} and XLM-Roberta \cite{xlm-roberta}, while the document pre-trained language models include LayoutXLM \cite{xfund}, LiLT \cite{lilt}, and GraphDoc \cite{graphdoc}, while the latter two models incorporate InfoXLM-base as the language model. 
As illustrated in Figure \ref{fig:SRFUND_exp_models}, for tasks 1, 2, and 4, we use a symmetric attention relation matrix as the learning target, where each aggregation target may contain multiple fine-grained element sets. Each aggregation target is considered an independent prediction sample, and a prediction is deemed correct only if all elements within the target set are completely aggregated together. For task 5, we utilize an asymmetric attention relation matrix to learn the relationships between different entities
, while a pair prediction is only considered correct if the directional relationship between two entities is accurately predicted. 
We employ the F1 score as the final evaluation metric for all tasks, and \textit{Merger} \cite{graphdcm} is adopted as the default relation classification head in the following sections.

All models were run on servers equipped with eight 48 GB A40 graphics cards with a batch size of 8. To ensure that different models achieved their optimal performance, we followed the training strategies for vision-only models as the original version in mmdetection \cite{mmdetection}, with specific settings for learning rate, optimizer, and the number of epoches detailed in the appendix. The text-only and multi-modal models uniformly utilized the Adam \cite{Adam} optimizer with $\left(\beta_1, \beta_2\right) = \left(0.9, 0.999\right)$ and underwent 200 training epoches, with an initial learning rate set at 5e-5. The training began with a linear warm-up during the first 10\% epochs, followed by a continuation under a linear decay strategy.



\subsection{Results and analysis}

\subsubsection{Word to text-line merging}
\label{w2l}

\begin{table}[h]
\caption{Results of the word to text-line merging task, using F1-score as the metric.}
\resizebox{\textwidth}{!}{
\begin{tabular}{clccccccccc}
\toprule
Type                         & Method       & English & Chinese & Japanese & German & French & Spanish & Italian & Portuguese & Avg. \\ \midrule
\multirow{3}{*}{Vision-only} & YOLOX \cite{yolox}        & 0.8222 & 0.8053 & 0.6959 & 0.8587 & 0.7310 & 0.8301 & 0.7470 & 0.7900 & 0.7850  \\ \cmidrule(l){2-11} 
                             & Cascade-RCNN \cite{cascadercnn} & 0.8520 & 0.8842 & 0.7569 & 0.8683 & 0.8191 & 0.8404 & 0.7590 & 0.7710 & 0.8189  \\ \cmidrule(l){2-11} 
                             & DAB-DETR \cite{dabdetr}     & 0.8437 & 0.8500 & 0.7394 & 0.8795 & 0.8082 & 0.8468 & 0.7926 & 0.7954 & 0.8194  \\ \midrule
\multirow{2}{*}{Text-only}   & XLM-RoBerta \cite{xlm-roberta}  & 0.6290 & 0.6272 & 0.6093 & 0.6982 & 0.6921 & 0.6470 & 0.6285 & 0.6780 & 0.6509  \\ \cmidrule(l){2-11} 
                             & InfoXLM \cite{infoxlm}      & 0.6426 & 0.6482 & 0.6298 & 0.7011 & 0.6974 & 0.6551 & 0.6253 & 0.6921 & 0.6611  \\ \midrule
\multirow{3}{*}{Multi-modal} & LayoutXLM \cite{layoutlm}    & \textbf{0.9081} & 0.9360 & \textbf{0.9118} & \textbf{0.9255} & \textbf{0.9282} & \textbf{0.9372} & \textbf{0.9157} & \textbf{0.9387} & \textbf{0.9260}  \\ \cmidrule(l){2-11} 
                             & LiLT \cite{lilt}         & 0.8887 & \textbf{0.9387} & 0.8803 & 0.9193 & 0.9223 & 0.9202 & 0.8962 & 0.9054 & 0.9094  \\ \cmidrule(l){2-11} 
                             & GraphDoc \cite{graphdoc}     & 0.8755 & 0.9100 & 0.8005 & 0.9167 & 0.8954 & 0.8993 & 0.8471 & 0.8708 & 0.8758  \\ \bottomrule
\end{tabular}
}
\label{tab:word-to-text}
\end{table}

\textbf{Aggregating words into text lines presents significant challenges for single-modal approaches.}  As demonstrated in Table \ref{tab:word-to-text}, text-only models faced difficulties due to the absence of two-dimensional spatial coordinates and visual cues.  This lack of information impedes their ability to accurately identify breakpoints in semantically continuous texts that span multiple lines.  Conversely, vision-only models depend exclusively on the visual boundary features of text lines to assess word aggregation.  This approach often fails to correctly segment words that are visually close yet semantically distinct, such as \textit{Question} and \textit{Answer} appearing on the same line. Additionally, the intricate form layouts of languages like Japanese and Italian pose further challenges in precise text-line segmentation.  In contrast, document pre-trained language models that integrate multiple modalities significantly improve performance by leveraging a broader range of data, thereby overcoming the limitations of single-modal systems.

\subsubsection{Text-line to entity merging}
\label{l2e}

\begin{table}[h]
\caption{Results of the text-line to entity merging task, using F1-score as the metric.}
\resizebox{\textwidth}{!}{
\begin{tabular}{clccccccccc}
\toprule
Type                     & Method       & English & Chinese & Japanese & German & French & Spanish & Italian & Portuguese & Avg. \\ \midrule
\multirow{3}{*}{Vision-only} & YOLOX \cite{yolox}        & 0.7415 & 0.7243 & 0.5891 & 0.7309 & 0.6504 & 0.7449 & 0.6238 & 0.6594 & 0.6830 \\ \cmidrule(l){2-11} 
                             & Cascade-RCNN \cite{cascadercnn} & 0.7918 & 0.8336 & 0.6873 & 0.7997 & 0.8060 & 0.8138 & 0.7153 & 0.7560 & 0.7754 \\ \cmidrule(l){2-11} 
                             & DAB-DETR \cite{dabdetr}     & 0.7681 & 0.7794 & 0.6332 & 0.7893 & 0.7344 & 0.7663 & 0.7075 & 0.7346 & 0.7391 \\ \midrule
\multirow{2}{*}{Text-only}   & XLM-RoBerta \cite{xlm-roberta}  & 0.8767 & 0.9354 & 0.8974 & 0.8850 & 0.9014 & 0.9044 & 0.8836 & 0.9226 & 0.9029 \\ \cmidrule(l){2-11} 
                             & InfoXLM \cite{infoxlm}      & 0.8773 & 0.9411 & 0.8921 & 0.8729 & 0.9010 & 0.9026 & 0.8847 & 0.9188 & 0.9012 \\ \midrule
\multirow{3}{*}{Multi-modal} & LayoutXLM \cite{layoutlm}    & 0.9151 & \textbf{0.9681} & \textbf{0.9387} & \textbf{0.9157} & \textbf{0.9408} & \textbf{0.9463} & \textbf{0.9280} & \textbf{0.9594} & \textbf{0.9412} \\ \cmidrule(l){2-11} 
                             & LiLT \cite{lilt}         & 0.9047 & 0.9542 & 0.9117 & 0.9140 & 0.9368 & 0.9351 & 0.9134 & 0.9430 & 0.9283 \\ \cmidrule(l){2-11} 
                             & GraphDoc \cite{graphdoc}     & \textbf{0.9229} & 0.9343 & 0.8770 & 0.9113 & 0.9260 & 0.9326 & 0.9060 & 0.9314 & 0.9181 \\ \bottomrule
\end{tabular}
}
\label{tab:line-to-entity}
\end{table}

\textbf{The task of merging text lines into entities relies more on semantic information.} Compared to task 1, we can observe from Table \ref{tab:line-to-entity} that the performance of text-only models surpasses that of vision-only models. This indicates that pure vision models have a weaker capability in understanding multi-line entities, while linguistic information significantly aids in capturing the semantic continuity between different text lines. 

\textbf{The pre-training process greatly impacts the effectiveness of document pre-trained language models.} The GraphDoc model, which leverages sentence-level semantic information and is only pre-trained on English documents, performs well in the task of merging text lines in English forms. In contrast, LayoutXLM is pre-trained using documents in all languages included in the SRFUND dataset, which allows it to demonstrate superior performance on forms in other languages.

\subsubsection{Entity category classification}
\label{el}

\begin{table}[h]
\caption{Results of the entity category classification task, using F1-score as the metric.}
\resizebox{\textwidth}{!}{
\begin{tabular}{clccccccccc}
\toprule
Type                     & Method       & English & Chinese & Japanese & German & French & Spanish & Italian & Portuguese & Avg. \\ \midrule
\multirow{3}{*}{Vision-only} & YOLOX \cite{yolox}        & 0.5284 & 0.6040 & 0.4619 & 0.4976 & 0.4743 & 0.5385 & 0.4466 & 0.4244 & 0.4969 \\ \cmidrule(l){2-11} 
                             & Cascade-RCNN \cite{cascadercnn} & 0.6739 & 0.7482 & 0.6124 & 0.7123 & 0.7749 & 0.7318 & 0.6662 & 0.6707 & 0.6988 \\ \cmidrule(l){2-11} 
                             & DAB-DETR \cite{dabdetr}     & 0.6531 & 0.6631 & 0.5286 & 0.6735 & 0.6863 & 0.6574 & 0.6067 & 0.6152 & 0.6355 \\ \midrule
\multirow{2}{*}{Text-only}   & XLM-RoBerta \cite{xlm-roberta}  & 0.8558 & 0.9666 & 0.8847 & 0.8912 & 0.9067 & 0.9161 & 0.8955 & 0.8884 & 0.9028  \\ \cmidrule(l){2-11} 
                             & InfoXLM \cite{infoxlm}      & 0.8589 & 0.9570 & 0.8782 & 0.8953 & 0.9107 & 0.9221 & 0.8995 & 0.8840 & 0.9025  \\ \midrule
\multirow{3}{*}{Multi-modal} & LayoutXLM \cite{layoutlm}    & \textbf{0.9045} & \textbf{0.9718} & \textbf{0.8957} & 0.9216 & \textbf{0.9299} & \textbf{0.9320} & \textbf{0.9269} & \textbf{0.9086} & \textbf{0.9248}  \\ \cmidrule(l){2-11} 
                             & LiLT \cite{lilt}         & 0.8678 & 0.9631 & 0.8876 & 0.9006 & 0.9217 & 0.9270 & 0.9135 & 0.8967 & 0.9118  \\ \cmidrule(l){2-11} 
                             & GraphDoc \cite{graphdoc}     & 0.8930 & 0.9619 & 0.8620 & \textbf{0.9261} & 0.9129 & 0.9250 & 0.9169 & 0.8897 & 0.9113  \\ \bottomrule
\end{tabular}
}
\label{tab:entity_classification}
\end{table}

\textbf{Visual modalities can also address the task of entity classification by learning layout information.} Apart from \textit{Header} entities, which typically exhibit characteristics such as boldface and larger font sizes, the variations in font styles among other entity categories are minimal. However, as shown in Table \ref{tab:entity_classification}, detection models are also capable of effectively handling the detection tasks for various entity categories. This indicates that the visual modality can acquire layout information, such as \textit{Question} typically being located to the left of \textit{Answer}, and entities with multi-line texts usually being categorized as \textit{Answer} or \textit{Other}.

\subsubsection{Item table localization}
\label{tableloc}

\begin{table}[h]
\caption{Results of the item table localization task, using F1-score as the metric.}
\resizebox{\textwidth}{!}{
\begin{tabular}{clccccccccc}
\toprule
Type                     & Method       & English & Chinese & Japanese & German & French & Spanish & Italian & Portuguese & Avg. \\ \midrule
\multirow{3}{*}{Vision-only} & YOLOX \cite{yolox}        & 0.1100 & 0.1721 & 0.0100 & 0.0467 & 0.1000 & 0.0710 & 0.0600 & 0.0911 & 0.0826 \\ \cmidrule(l){2-11} 
                             & Cascade-RCNN \cite{cascadercnn} & 0.0839 & 0.2081 & 0.0433 & 0.0800 & 0.1327 & \textbf{0.1427} & 0.0817 & \textbf{0.1486} & 0.1151 \\ \cmidrule(l){2-11} 
                             & DAB-DETR \cite{dabdetr}     & 0.1399 & 0.2670 & 0.0000 & 0.0667 & 0.1333 & 0.0903 & 0.0767 & 0.1100 & 0.1105 \\ \midrule
\multirow{2}{*}{Text-only}   & XLM-RoBerta \cite{xlm-roberta}  & 0.0526 & 0.2090 & 0.0800 & \textbf{0.5714} & 0.2222 & 0.0000 & 0.1333 & 0.0526 & 0.1514  \\ \cmidrule(l){2-11} 
                             & InfoXLM \cite{infoxlm}      & 0.0513 & 0.1846 & 0.0000 & 0.4545 & 0.2143 & 0.0000 & 0.0000 & 0.0000 & 0.1124  \\ \midrule
\multirow{3}{*}{Multi-modal} & LayoutXLM \cite{layoutlm}    & \textbf{0.7273} & \textbf{0.3333} & \textbf{0.1053} & 0.4348 & 0.1053 & 0.0588 & \textbf{0.3158} & 0.1250 & \textbf{0.3022}  \\ \cmidrule(l){2-11} 
                             & LiLT \cite{lilt}         & 0.2273 & 0.1867 & 0.0000 & 0.5263 & 0.0769 & 0.0000 & 0.0000 & 0.0417 & 0.1306  \\ \cmidrule(l){2-11} 
                             & GraphDoc \cite{graphdoc}     & 0.0000 & 0.0556 & 0.0000 & 0.3333 & \textbf{0.3333} & 0.0606 & 0.0000 & 0.1224 & 0.0945  \\ \bottomrule
\end{tabular}
}
\label{tab:table_localization}
\end{table}

\textbf{The item table localization task presents significant challenges.} Since this task requires that all entities within the item table be included, any missing entity predictions result in the outcome being deemed incomplete. This requirement makes it difficult for models of any modality to accurately locate item tables within documents, as shown in Table \ref{tab:table_localization}. 

\textbf{There is significant variability in performance across forms in different languages.} Item table localization in forms of different languages requires optimization through models of various modalities. For some languages, the localization of item tables relies more heavily on the capabilities of vision models, such as in Spanish and Portuguese forms.

\subsubsection{Hierarchical structure recovery}
\label{structure}

\begin{table}[h]
\caption{Results of the hierarchical structure recovery task, using F1-score as the metric.}
\resizebox{\textwidth}{!}{
\begin{tabular}{clccccccccc}
\toprule
Type                     & Method       & English & Chinese & Japanese & German & French & Spanish & Italian & Portuguese & Avg. \\ \midrule
\multirow{2}{*}{Text-only}   & XLM-RoBerta \cite{xlm-roberta}  & 0.5270 & 0.6514 & 0.5388 & 0.6637 & 0.6054 & 0.6121 & 0.5839 & 0.5081 & 0.5830   \\ \cmidrule(l){2-11} 
                             & InfoXLM \cite{infoxlm}      & 0.5305 & 0.6436 & 0.5227 & 0.6695 & 0.6071 & 0.5941 & 0.5736 & 0.4872 & 0.5732   \\ \midrule
\multirow{3}{*}{Multi-modal} & LayoutXLM \cite{layoutlm}    & 0.7135 & 0.7601 & 0.6626 & 0.7734 & 0.7415 & 0.7009 & 0.6710 & 0.6310 & 0.7013   \\ \cmidrule(l){2-11} 
                             & LiLT \cite{lilt}         & 0.7050 & 0.7578 & 0.6538 & 0.7499 & 0.7153 & 0.6940 & 0.6702 & 0.5747 & 0.6821   \\ \cmidrule(l){2-11} 
                             & GraphDoc \cite{graphdoc}     & \textbf{0.7938} & \textbf{0.7881} & \textbf{0.6714} & \textbf{0.7976} & \textbf{0.7754} & \textbf{0.7416} & \textbf{0.6969} & \textbf{0.6648} & \textbf{0.7349}   \\ \bottomrule
\end{tabular}
}
\label{tab:structure}
\end{table}


\textbf{The granularity of semantic information must align with the task requirements.} Among the three document pre-trained language models, only the GraphDoc model maintains sentence-level or entity-level linguistic input during both the pre-training stage and the training/testing phases of the current task. This alignment enables the GraphDoc model to achieve the best performance in the task of entity-based document structure recovery.

\subsubsection{Overall analysis}
The results from the aforementioned five tasks demonstrate that uni-modal models exhibit relatively poor performance, while document pre-trained models that incorporate multiple input modalities perform significantly better. Concurrently, no single model has consistently outperformed others across all tasks and languages, indicating that in practical applications, we cannot simply rely on a single model or approach to handle all types of form structuring tasks. Instead, we need to select appropriate models and strategies based on the specific requirements of the task and the characteristics of the language involved. This finding underscores the importance of adopting a nuanced and tailored approach when tackling form structuring challenges, rather than employing a one-size-fits-all solution.

\section{Conclusion}
In summary, this paper presents two main contributions. Firstly, we propose a multilingual, multitask document hierarchical structuring benchmark named SRFUND, encompassing 1,592 forms from eight languages. To the best of our knowledge, this is the first benchmark in form understanding that integrates multi-level structure reconstruction, spanning from words to the global structure of forms. Secondly, we conducted baseline experiments on five tasks using various representative approaches from different modalities, demonstrating that the SRFUND benchmark introduces new challenges to the field of form understanding. We believe that the SRFUND benchmark holds significant potential for future academic research, contributing continuously to the in-depth study of global form structures.

\bibliographystyle{plain}

\bibliography{neurips_data_2024}

\clearpage
\newpage

\section*{Checklist}


\begin{enumerate}

\item For all authors...
\begin{enumerate}
  \item Do the main claims made in the abstract and introduction accurately reflect the paper's contributions and scope?
    \answerYes{}
  \item Did you describe the limitations of your work?
    \answerNo{}
  \item Did you discuss any potential negative societal impacts of your work?
    \answerNo{}
  \item Have you read the ethics review guidelines and ensured that your paper conforms to them?
    \answerYes{}
\end{enumerate}

\item If you are including theoretical results...
\begin{enumerate}
  \item Did you state the full set of assumptions of all theoretical results?
    \answerNA{}We don’t have theoretical results.
	\item Did you include complete proofs of all theoretical results?
    \answerNA{}We don’t have theorems.
\end{enumerate}

\item If you ran experiments (e.g. for benchmarks)...
\begin{enumerate}
  \item Did you include the code, data, and instructions needed to reproduce the main experimental results (either in the supplemental material or as a URL)?
    \answerYes{}As a URL
  \item Did you specify all the training details (e.g., data splits, hyperparameters, how they were chosen)?
    \answerYes{}In Sec. \ref{sec:Experiments}
	\item Did you report error bars (e.g., with respect to the random seed after running experiments multiple times)?
    \answerNo{}
	\item Did you include the total amount of compute and the type of resources used (e.g., type of GPUs, internal cluster, or cloud provider)?
    \answerYes{}In Sec. \ref{sec:Experiments}
\end{enumerate}

\item If you are using existing assets (e.g., code, data, models) or curating/releasing new assets...
\begin{enumerate}
  \item If your work uses existing assets, did you cite the creators?
    \answerYes{}
  \item Did you mention the license of the assets?
    \answerYes{}
  \item Did you include any new assets either in the supplemental material or as a URL?
    \answerNA{}
  \item Did you discuss whether and how consent was obtained from people whose data you're using/curating?
    \answerNA{}
  \item Did you discuss whether the data you are using/curating contains personally identifiable information or offensive content?
    \answerNA{}
\end{enumerate}

\item If you used crowdsourcing or conducted research with human subjects...
\begin{enumerate}
  \item Did you include the full text of instructions given to participants and screenshots, if applicable?
    \answerNA{}
  \item Did you describe any potential participant risks, with links to Institutional Review Board (IRB) approvals, if applicable?
    \answerNA{}
  \item Did you include the estimated hourly wage paid to participants and the total amount spent on participant compensation?
    \answerNA{}
\end{enumerate}

\end{enumerate}


\end{document}